%% file: main.tex
\documentclass{article}
\usepackage{ijcai23}

\usepackage{times}
\usepackage{soul}
\usepackage{url}
\usepackage[hidelinks]{hyperref}
\usepackage[utf8]{inputenc}
\usepackage[small]{caption}
\usepackage{graphicx}
\usepackage{amsmath}
\usepackage{amsthm}
\usepackage{booktabs}
\usepackage{algorithm}
\usepackage{algorithmic}
\usepackage[switch]{lineno}

\usepackage[table]{xcolor}
\usepackage{multirow}
\usepackage{multicol}
\usepackage{amsmath}
\usepackage{amssymb}

\title{
Graph Sampling-based Meta-Learning for Molecular Property Prediction
}

\author{
Xiang Zhuang$^{1,2,3}$\thanks{Equal contribution and shared co-first authorship.}
\and
Qiang Zhang$^{1,2}$\footnotemark[1]\thanks{Corresponding author.}
\and
Bin Wu$^{2}$\and
Keyan Ding$^{2}$\and
Yin Fang$^{1,2,3}$\And 
\\
Huajun Chen$^{1,2,3}$\footnotemark[2]
\affiliations
$^1$College of Computer Science and Technology, Zhejiang University\\
$^2$ZJU-Hangzhou Global Scientific and Technological Innovation Center\\
$^3$Alibaba-Zhejiang University Joint Research Institute of Frontier Technologies
\emails
\{zhuangxiang,qiang.zhang.cs,binwu,dingkeyan,fangyin,huajunsir\}@zju.edu.cn
\\
}

\begin{document}

\maketitle

\begin{abstract}

Molecular property is usually observed with a limited number of samples, and researchers have considered property prediction as a few-shot problem. 
One important fact that has been ignored by prior works is that each molecule can be recorded with several different properties simultaneously.  
To effectively utilize many-to-many correlations of molecules and properties, we propose a Graph Sampling-based Meta-learning (GS-Meta) framework {for few-shot molecular property prediction}.
First, we construct a Molecule-Property relation Graph (MPG): molecule and properties are nodes, while property labels decide edges. 
Then, to utilize the topological information of MPG,   
we reformulate an episode in meta-learning as a subgraph of the MPG, containing a target property node, molecule nodes, and auxiliary property nodes.
Third, as episodes in the form of subgraphs are no longer independent of each other, {we propose to schedule the subgraph sampling process with {a contrastive loss function}, which considers the consistency and discrimination of subgraphs.}
Extensive experiments on 5 commonly-used benchmarks show GS-Meta consistently outperforms state-of-the-art methods by 
{5.71}\%-{6.93}\% {in ROC-AUC} and verify the effectiveness of each proposed module. Our code is available at \url{https://github.com/HICAI-ZJU/GS-Meta}.
\end{abstract}

\section{Introduction}
Drug discovery is of great significance to public health and the development of new drugs is a long and costly process. In the early lead optimization phase, researchers need to select a large number of molecules as candidates and conduct virtual screening to avoid wasting resources on molecules that are unlikely to possess the desired properties~\cite{riniker2013similarity,sliwoski2014computational}. Recently, deep learning plays an important role in this process, and several deep models have been investigated to predict molecular property~\cite{DBLP:conf/ijcai/SongZNFLY20,DBLP:conf/aaai/FangZYZD0Q0FC22,fang2023knowledge}. In the practical settings, only a few molecules can be made and tested in the wet-lab experiments~\cite{altae2017low,DBLP:conf/www/GuoZYHW0C21}. The limited amount of annotated data often hinders the generalization ability of deep learning models in practical applications \cite{DBLP:conf/nips/BertinettoHVTV16}.

\begin{figure}[!t]
\centering
\includegraphics[width=0.9\columnwidth]{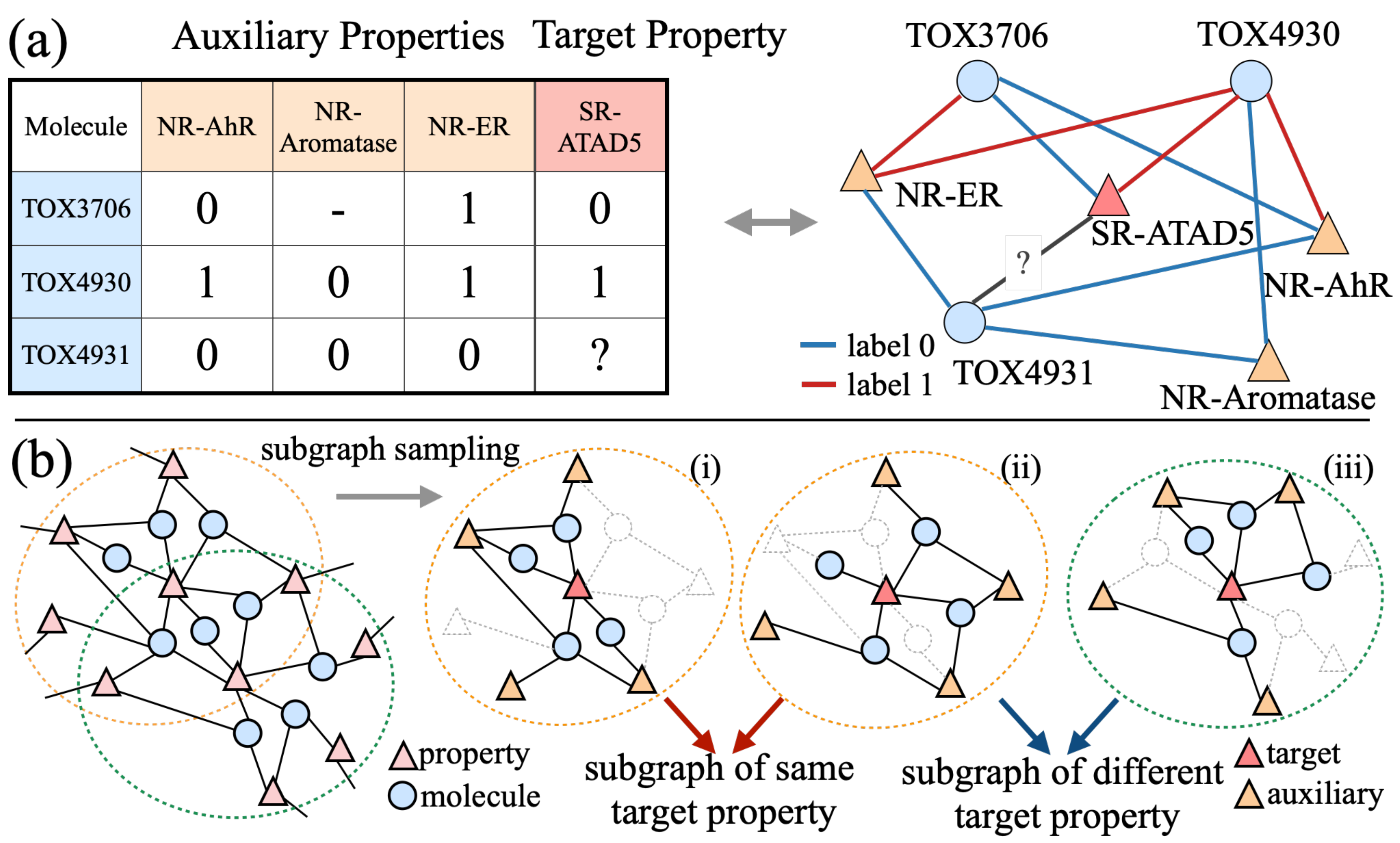} 

\caption{(a) An example of the Molecule-Property relation Graph (MPG) built based on the Tox21 dataset \protect\cite{wu2018moleculenet}. To predict the SR-ATAD5 property of the TOX4931 molecule, we can utilize some auxiliary properties to form an MPG where molecules and properties are nodes and edges represent property labels (0 or 1). (b) An example of subgraph sampling. Given the MPG, we can sample a pair of subgraphs with the same target property (\textit{e.g.}, (i) and (ii)) as positive samples and another pair with different target properties (\textit{e.g.}, (ii) and (iii)) as negative samples.}
\label{fig:intro}
\end{figure}

Deficient annotated data is often termed a few-shot learning (FSL) problem~\cite{DBLP:journals/csur/WangYKN20}. Typically, meta-learning~\cite{hospedales2021meta}, which aims to learn how to learn, is used to solve the few-shot problem and several prior works have incorporated meta-learning into molecular property prediction. For example, Meta-GNN~\cite{DBLP:conf/www/GuoZYHW0C21} {employed} a classic meta-learning method, MAML~\cite{DBLP:conf/icml/FinnAL17}, with self-supervised tasks including bond reconstruction and atom type prediction.~\citeauthor{DBLP:conf/nips/WangAYD21} {constructed} a relation graph of molecules and designed a meta-learning strategy to selectively update model parameters.

However, these works have neglected an important fact that, unlike common FSL settings including image classification, a molecule can be observed with multiple properties simultaneously, 
{such as a number of possible side effects.}
There is also a correlation between different properties, for example, endocrine disorders and cardiac diseases caused by drug side effects may occur together because they share disease pathway~\cite{fuchs2020high}. In predicting the property of a molecule, we can leverage other available labeled properties of the same molecule. When we know some properties of {one molecule} and are faced with new properties that have fewer labels, {we postulate that} utilizing these known property labels can alleviate the {label insufficiency} problem.

To effectively utilize such correlations, we propose a Graph Sampling-based Meta-learning framework, GS-Meta.
First, to accurately describe the many-to-many relations among molecules and properties, we build a Molecule-Property relation Graph (MPG), where nodes are molecules and properties, and edges between molecules and properties indicate the label of the molecules in the properties (Figure~\ref{fig:intro}(a)). 
Second, to employ the inherent graph topology of MPG, we propose to reformulate an episode as a subgraph of MPG, {which is composed of} a target property node, molecule nodes as well as auxiliary property nodes (Figure~\ref{fig:intro}(b)). 
Third, in conventional meta-learning \cite{DBLP:conf/nips/VinyalsBLKW16}, episodes are considered to be independently distributed and sampled from a uniform distribution. However, subgraphs are connected in our MPG due to intersecting molecules and edges, they are no longer independent of each other. 
We propose {a learnable sampling scheduler} and specify the subgraph dependency in two aspects: (1) subgraphs centered on the same target property node can be seen as different views describing the same task and thus they should be consistent with each other; and (2) subgraphs centered on
different target property nodes are episodes of different tasks,
and their semantic discrepancy should be enlarged. Hence, we {solve} this dependency via {a contrastive loss function}. In short, our contributions are summarized as follows: 
\begin{itemize}
    \item We propose to use auxiliary properties when facing new target property in the few-shot regime and construct a Molecule-Property relation Graph to model the relations among molecules and properties so that the information of the relevant properties can be used through the topology of the constructed graph.
    \item We propose a Graph Sampling-based Meta-learning framework, which reformulates episodes in meta-learning as sampled subgraphs from the constructed MPG, {and schedules the subgraph sampling process with {a contrastive loss function}.}
    \item Experiments on five benchmarks show that our method consistently outperforms state-of-the-art FSL molecular property prediction methods by {5.71}\%-{6.93}\% {in terms of ROC-AUC}.
\end{itemize}

\section{Related Work}
\paragraph{Few-shot Learning for Molecules}
The few-shot learning (FSL) problem~\cite{DBLP:conf/nips/VinyalsBLKW16,DBLP:conf/iclr/ChenLKWH19} occurs when there are limited labeled training data per object of interest. Often, meta-learning is used to solve the few-shot learning problem.
For example, MAML~\cite{DBLP:conf/icml/FinnAL17} learns a good parameter initialization and updates through gradient descents. {However, the existing methods are usually investigated for image classification but not tailored to the different settings of molecular property prediction~\cite{DBLP:conf/nips/WangAYD21}.} Recent efforts have been paid to fill this gap. \citeauthor{DBLP:conf/www/GuoZYHW0C21} propose to use molecule-specific tasks including masked atoms and bonds prediction to guide the model focus on the intrinsic characteristics of molecules. {\citeauthor{DBLP:conf/nips/WangAYD21} connect molecules in a homogeneous graph to propagate limited information between similar instances.} However, all the prior works have ignored the relationships between molecular properties, \textit{i.e.}, some auxiliary available properties can be used in predicting new molecular properties, and they fail to investigate the relationship.

\paragraph{Episode Scheduler in Meta-learning}
Most meta-learning approaches leverage a uniform episode sampling strategy in the training process. To exploit relations between episodes, prior methods have investigated how to schedule episodes to enhance the generalization of meta-knowledge. \citeauthor{DBLP:conf/eccv/LiuWSFZH20} design a greedy class-pair based strategy rather than uniform sampling. \citeauthor{DBLP:conf/iclr/FeiLXH21} consider the relationship of episodes to overcome the poor sampling problem. \citeauthor{DBLP:conf/nips/YaoWWZMLF21} propose a neural scheduler to decide which tasks to select from a candidate set. {In this work, we schedule episodes with the lens of subgraph sampling and encourage the consistency between subgraphs of the same target property and discrimination between different target properties via a contrastive loss function.

}

\begin{figure*}[!ht]
\centering
\includegraphics[width=0.90\textwidth]{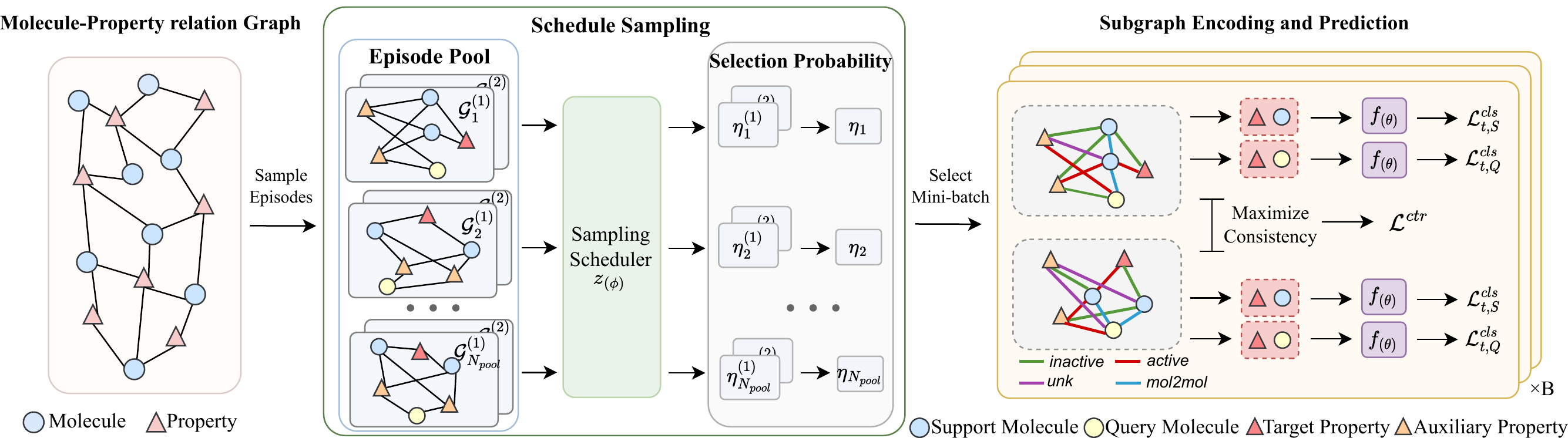} 
\caption{A 2-way 1-shot overview of GS-Meta. Firstly a Molecule-Property relation Graph (MPG) is constructed. Then, an episode candidate pool is randomly sampled, and the subgraph sampling scheduler $z_{(\phi)}$ is used to compute the selection probability $\eta$ and select a mini-batch containing $B$ episode pairs from the candidate pool. 
Finally, the relation learning module $f_{(\theta)}$ encodes each subgraph and outputs a classification loss $\mathcal{L}^{cls}$ within an episode and a contrastive loss $\mathcal{L}^{ctr}$ across episodes.}
\label{fig:framework}
\end{figure*}

\section{Graph Sampling-based Meta-Learning}
This section presents the proposed Graph Sampling-based Meta-learning framework, as shown in {Figure~\ref{fig:framework}}. We first define the few-shot molecular property prediction problem (Section~\ref{sec:problem-def}). To establish and exploit the many-to-many relation between molecules and properties, we construct a Molecule-Property relation Graph (MPG) and then reformulate an episode in meta-learning as a subgraph from the MPG (Section~\ref{sec:graph}). With this reformulation, we investigate to consider consistency and discrimination between subgraphs and schedule the subgraph sampling to facilitate meta-learning (Section~\ref{sec:schedule}). Finally, the training and testing strategies are described (Section~\ref{sec:process}).

\subsection{Problem Definition}
\label{sec:problem-def}
{Following \cite{altae2017low,DBLP:conf/www/GuoZYHW0C21}, the few-shot molecular property prediction is conducted on a set of tasks $\left\{\mathcal{T}\right\}$, {where each task $\mathcal{T}$ is to predict a property $p$ of molecules}. The training set $\mathcal{D}_{train}$ consists of several tasks $\left\{\mathcal{T}_{train}\right\}$, denoted as $\mathcal{D}_{train}=\left\{(x_i, y_{i,t})|t \in \mathcal{T}_{train}\right\}$, where $x_i$ is a molecule and $y_{i,t}$ is the label of $x_i$ on the $t$-th task. And the testing set $\mathcal{D}_{test}=\left\{(x_j, y_{j,t})|t \in \mathcal{T}_{test}\right\}$ is composed of a set of tasks $\left\{\mathcal{T}_{test}\right\}$. $\{p_{train}\}$ and $\{p_{test}\}$ are denoted as properties corresponding to tasks $\{\mathcal{T}_{train}\}$ and $\{\mathcal{T}_{test}\}$, and
training properties and testing properties are disjoint, \textit{i.e.}, $\{p_{train}\} \cap \{p_{test}\}=\emptyset$. The objective is to learn a predictor on $\mathcal{D}_{train}$ and to predict {novel} properties with a few labeled molecules in $\mathcal{D}_{test}$.}

To deal with the few-shot problem, the episodic training paradigm has shown great promise in meta-learning~\cite{DBLP:conf/icml/FinnAL17}. {Without loading all training tasks $\{\mathcal{T}_{train}\}$ into memory, batches of episodes $\{E_t\}_{t=1}^{B}$ are sampled iteratively in practical training process.} To construct an episode $E_t$,
a target task $\mathcal{T}_t$ is firstly sampled from $\left\{\mathcal{T}_{train}\right\}$, then a labeled support set $S_t$ and a query set $Q_t$ are sampled.
Usually, there are two classes {(\textit{i.e.}, \textit{active} ($y$=1) or \textit{inactive} ($y$=0))} in each molecular property prediction task,
and a 2-way \textit{K}-shot episode means that the support set $\mathcal{S}_t$ consists of 2 classes with \textit{K} molecules per class, \textit{i.e.}, $\mathcal{S}_t=\left\{\left(x_{i}^s, y_{i,t}^s\right)\right\}_{i=1}^{2 K}$, and query set $\mathcal{Q}_t=\left\{\left(x_{i}^q, y_{i,t}^q\right)\right\}_{i=1}^{M}$ contains $M$ molecules. In this case, we can define the episode as $E_t=(\mathcal{S}_t,\mathcal{Q}_t)$.

\subsection{Molecule-Property Relation Graph}
\label{sec:graph}
\paragraph{Graph Construction and Initialization}
To leverage the rich information behind the relations among properties and molecules, we construct a Molecule-Property relation Graph (MPG) that explicitly describes such relations. The graph is denoted as $\mathcal{G}=(\mathcal{V}, \mathcal{E})$, where $\mathcal{V}$ denotes the node set and $\mathcal{E}$ denotes the set of edges $e \in \mathcal{E}$. And there are two types of nodes in the graph, \textit{i.e.}, $\mathcal{V}=\mathcal{V}_{M} \cup \mathcal{V}_{T}$, where $\mathcal{V}_M=\left\{x_{m}\right\}$ is the molecule node set and $\mathcal{V}_T=\left\{p_{t}\right\}$ is the property node set. Edges $\mathcal{E}\subseteq\mathcal{V}_M\times\mathcal{V}_T$ are connected between these two types of nodes and the edge type of $e_{i,j}$ is initialized according to the label $y_{i,j}$, {\textit{i.e.}, \textit{active} (for $y$=1) or \textit{inactive} (for $y$=0)}.

For a molecule $x_i$, a graph-based encoder~\cite{DBLP:conf/iclr/XuHLJ19} is used following ~\cite{DBLP:conf/www/GuoZYHW0C21} to obtain its embedding:
\begin{equation}
    \boldsymbol{x}_{i}=f_{mol}(x_i),
\end{equation}
where $\boldsymbol{x}_{i} \in \mathbb{R}^d$. For a property node $p_t$, its embedding is randomly initialized {for simplicity} with the same length as molecules, \textit{i.e.}, $\boldsymbol{p}_t\in\mathbb{R}^d$. Hence the node features $\boldsymbol{h}_i^0$ are initialized with molecule and property embeddings respectively:
\begin{equation}    \boldsymbol{h}_i^0=\begin{cases}\boldsymbol{p}_i & \text{for } i \in \mathcal{V}_T
    \\
    \boldsymbol{x}_i & \text{for } i \in \mathcal{V}_M\end{cases}.
\end{equation}

Moreover, due to the deficiency of data, some molecules may not have labels on some auxiliary properties, leading to a missing edge connection between these molecules and properties. To make the graph topology compact, a special edge type \textit{unk} is used to complement the missing label.

\paragraph{Reformulating Episode as Subgraph}

The constructed MPG can be very large, \textit{e.g.} in the PCBA dataset \cite{wang2012pubchem}, there are more than 430,000 molecules and 128 properties, {and the corresponding MPG can contain more than 10 million edges. It is therefore computationally infeasible to directly work on the whole graph to predict molecular properties. Instead, we resort to subgraphs sampled from the MPG and connect them to episodes in meta-learning.}

{The episodic meta-learning, which is trained iteratively by sampling batches of episodes, has proven to be an effective training strategy~\cite{DBLP:conf/icml/FinnAL17}. Therefore, to adopt episodic meta-learning on MPG, we propose to reformulate the episode as a subgraph.}
Specifically, an episode of task $\mathcal{T}_t$ is equivalent to a subgraph containing a property node $p_t$ in $\mathcal{V}_T$, and molecules connected to $p_t$.
{Hence}, the support set $\mathcal{S}_t$ can be reformulated as a subgraph of MPG:
\begin{equation}
\mathcal{S}_t \sim \mathcal{G}^S_t =\left\{\left(x_i,e_{i,t},p_t\right),y_{i,t}|x_i\in\mathcal{N}(p_t)\right\}_{i=1}^{2K},
\end{equation}
where $p_t$ is a property node, and $\mathcal{N}(p_t)$ is the neighbors of $p_t$. A query molecule is sampled as the query set, denoted as:
\begin{equation}
    \mathcal{Q}_t\sim \mathcal{G}^Q_t=\{(x_j, p_t), y_{j,t}|x_j\in\mathcal{N}(p_t)\}.
\end{equation}
By merging two subgraphs, the episode $E_t$ is reformulated as:
\begin{equation}
E_t \sim \mathcal{G}_t=\mathcal{G}^S_t \cup \mathcal{G}^Q_t.
\end{equation}
The node and edge set of $\mathcal{G}_t$ are denoted as $\mathcal{V}^t$ and $\mathcal{E}^t$ respectively.

Since molecules have multiple properties, other available properties can be used when predicting a novel property of the same molecule. To leverage these auxiliary properties, we add some other property nodes connected to molecule nodes into the subgraph:
\begin{equation}
\mathcal{G}^A_t=\left\{\left(x_i,e_{i,a},p_a\right),y_{i,a}|p_a\in\mathcal{N}(x_i)\backslash p_t\right\}_{a=1}^{N_a},
\end{equation}
where $N_a$ is the number of auxiliary property, $p_a$ is the auxiliary property node, $x_i \in \mathcal{V}^t_M$ is a molecule node in $\mathcal{G}_t$, and $\mathcal{G}_t^A$ is an auxiliary subgraph. With $\mathcal{G}_t^A$, we extend $\mathcal{G}_t$ as:
\begin{equation}
\mathcal{G}_t=\mathcal{G}_t \cup \mathcal{G}_t^A.
\end{equation}
and there are totally $2K+N_a+2$ nodes, $2K$ support molecules, one query molecule, one target property, and $N_a$ auxiliary properties. \textbf{From here on out, an episode of meta-learning and a subgraph of the MPG are semantically equivalent in this paper.}

\paragraph{Subgraph Encoding and Prediction}
We adopt a massage passing schema~\cite{DBLP:conf/nips/HamiltonYL17} to iteratively update each sampled subgraph $\mathcal{G}_t$. {In contrast to the previous work which only takes labels as edges~\cite{cao2021relational}},
we also consider relations between molecules, and design an edge predictor to estimate connections between molecules at each iteration:
\begin{equation}
\label{eq:edge-pred}
\alpha_{i,j}^l=\sigma\left(\mathrm{MLP}\left(\exp\left(-|{\boldsymbol{h}_i^{l-1}}-\boldsymbol{h}_j^{l-1}|\right)\right)\right),
\end{equation}
where $\sigma$ is Sigmoid function, $\boldsymbol{h}_i^{l-1}$ and $\boldsymbol{h}_j^{l-1}$ are embeddings of molecules $x_i$ and $x_j$ at ($l$-1)-th iteration respectively, and $\alpha_{i,j}^l$ is the estimated connection weight between $x_i$ and $x_j$. To avoid connecting {dissimilar molecules}, only top-\textit{k} ({\textit{k} is hyper-parameter}) edges with the largest estimated weights are kept and the edge type is \textit{mol2mol}. Overall, there are four types of edges in $\mathcal{G}_t$, which are \textit{active}, \textit{inactive} and \textit{unk} between a molecule and a property, and \textit{mol2mol} between molecules. 

After constructing the complete subgraph, node embeddings are updated as follows:
\begin{equation}
\label{eq:message}
    \boldsymbol{h}_i^l=\mathrm{GNN}^{l-1}\left(\boldsymbol{h}_i^{l-1},\boldsymbol{h}_j^{l-1},\boldsymbol{h}_{i,j}^{l-1},w_{i,j}^l|j\in\mathcal{N}\left(i\right)\right),
\end{equation}
where $\boldsymbol{h}_i^l$ is the embedding of node $i$ at the \textit{l}-th iteration, $\boldsymbol{h}_{i,j}^l$ is edge embedding initialized according to edge type, $\mathcal{N}(i)$ is neighbors of node $i$, and $w_{i,j}^l$ is the edge weight defined as:
\begin{equation}
    w_{i,j}^l=\begin{cases}\alpha_{i,j}^l & e_{i,j}\in\mathcal{E}_{M,M} \\ 1 & \text{otherwise}\end{cases},
\end{equation}
where $\mathcal{E}_{M,M}$ is the set of edges between molecules. After $L$ iterations, the final embedding of molecule $\boldsymbol{h}_i^L$ and the property $\boldsymbol{h}_t^L$ are we concatenated to predict the label:
\begin{equation}
\label{eq:classfier}
    \hat{y}_{i,t}=\sigma\left(f_{cls}\left([\boldsymbol{h}_i^L\oplus \boldsymbol{h}_t^L]\right)\right),
\end{equation}
where $\hat{y}_{i,t} \in \mathbb{R}^{1}$ is the prediction, and $\oplus$ is a concatenation operation. \textbf{For simplicity, we denote $f_{(\theta)}$ as the relation learning module with parameter $\theta$, which includes molecular encoder $f_{mol}$, property initial embeddings, GNN layers, edge predictors, and classifier $f_{cls}$}. More details are illustrated in Appendix~\ref{sec:msg-detail}.

In each subgraph $\mathcal{G}_t$, the task classification loss on the support set $\mathcal{S}_t$ is calculated:
\begin{equation}
\label{eq:cls_ts}
\mathcal{L}^{cls}_{t,\mathcal{S}}\left(f_{(\theta)}\right)=-\textstyle \sum_{\mathcal{S}_t}\left(y \log \hat{y}+\left(1-y\right) \log \left(1-\hat{y}\right)\right),
\end{equation}
where {$y$ and $\hat{y}$ are short for $y_{i,t}$ and $\hat{y}_{i,t}$} for clarity. Similarly, we can calculate the classification loss on the query set $\mathcal{L}^{cls}_{t,\mathcal{Q}}$.

\subsection{Subgraph Sampling Scheduler}
\label{sec:schedule}
{Previous few-shot molecular property prediction methods~\cite{DBLP:conf/www/GuoZYHW0C21,DBLP:conf/nips/WangAYD21}} randomly sample episodes with a uniform probability, under the assumption that they are independent and of equal importance.
{However, subgraphs centered on different target properties are potentially connected to each other in the built MPG, due to the existence of intersecting nodes or edges. Hence, considering the subgraph dependency, we are motivated to develop a subgraph sampling scheduler that determines which episodes to use in the current batch during meta-training.
}

\paragraph{Consistency and Discrimination between Subgraphs}

We specify the subgraph dependency in two aspects.
(1) Each subgraph only has a small number of molecules ($K$ per class), and cannot contain all the information about the target property. 
For the same target property, subgraphs with different molecules (subgraph (i) and (ii) in Figure~\ref{fig:intro}(b)) can be seen as different views describing the same task and they should be consistent with each other.
(2) Meanwhile, subgraphs centered on different target property nodes (subgraph (ii) and (iii) in Figure~\ref{fig:intro}(b)) are episodes of different tasks, and their semantic discrepancy should be enlarged.

Towards this end, the subgraph sampling scheduler, denoted as $z_{(\phi)}$ with parameters $\phi$, adopts a pairwise sampling strategy. That is, two subgraphs $\mathcal{G}_t^{(1)}$ and $\mathcal{G}_t^{(2)}$ of the same target property $p_t$ are sampled simultaneously.
Specifically, at the beginning of each mini-batch, we randomly sample a pool of subgraph candidates, 
$\mathrm{P}=\{(\mathcal{G}_t^{(1)},\mathcal{G}_t^{(2)})\}_{t=1}^{N_{pool}}$, 
which are pairs of subgraphs for the same target property. 
For a subgraph $\mathcal{G}_t$, the scheduler $z_{(\phi)}$ outputs its selection probability $\eta_t$ via two steps. The first step calculates the subgraph embedding $\boldsymbol{g}_t$:
\begin{equation}
  \boldsymbol{g}_t = \boldsymbol{h}_t^L+ \sigma \left(\textstyle{\sum_{i \in \mathcal{V}^t \backslash p_t }  \boldsymbol{h}_i^L}\right),
\end{equation}
which is the pooling of the final embedding of each node, and $\boldsymbol{h}_t^L$ is final embedding of target property $p_t$. Then, we take $\boldsymbol{g}_t$ as input to the scheduler $z_{(\phi)}$ to get the selection probability $\eta_t$:
\begin{equation}
\label{eq:select}
\eta_t=z_1\left(\boldsymbol{g}_t+z_2\left(\textstyle{\sum_{t^\prime\in\mathrm{P}\backslash \mathcal{G}_t}\boldsymbol{g}_{t^\prime}}\right)\right),
\end{equation}
where $z_1$ and $z_2$ are MLP and together constitute $z_{(\phi)}$. Then, $\eta_t$ is normalized by softmax to get a reasonable probability value. Thus, for each episode pair $(\mathcal{G}_t^{(1)},\mathcal{G}_t^{(2)})$ in the candidate pool, the selection probability can be computed as $(\eta_t^{(1)}+\eta_t^{(2)})/2$, and we sample $B$ from $N_{pool}$ in the candidate pool to form a mini-batch according to selection probability. 

To encourage consistency between subgraphs of the same target property and discrimination between different target properties, we adopt the NT-Xent loss~\cite{DBLP:conf/iclr/HjelmFLGBTB19} which is widely used in contrastive learning. Subgraphs of the same target property are positive pairs and those of different target properties are negatives,
and the contrastive loss in a mini-batch is as follows:
\begin{equation}
\label{eq:contr}
    \mathcal{L}^{ctr}=\frac{1}{B}\sum_{t=1}^{B}-\log\frac{e^{\mathrm{sim}\left(\boldsymbol{g}_t^{(1)},\boldsymbol{g}_t^{(2)}\right)/\tau}}{\sum_{t^{\prime}=1,t^\prime\ne t}^{B}e^{\mathrm{sim}\left(\boldsymbol{g}_t^{(1)},\boldsymbol{g}_{t^\prime}^{(2)}\right)/\tau}},
\end{equation}
where $B$ is mini-batch size, $\mathrm{sim}(\cdot,\cdot)$ is cosine similarity and $\tau$ is the temperature parameter.

\begin{algorithm}[!t]
\caption{Training and optimization algorithm.}
\label{alg:algorithm}
\textbf{Input}: Molecule-Property relation Graph (MPG)\\
\textbf{Output}: Relation learning module $f_{(\theta)}$, subgraph sampling scheduler $z_{(\phi)}$.
\begin{algorithmic}[1] 
\WHILE{not done}
\STATE Sample $N_{pool}$ episode pairs $\{(\mathcal{G}_t^{(1)},\mathcal{G}_t^{(2)})\}_{t=1}^{N_{pool}}$ as candidates
\STATE Calculate selection probability $\eta_t$ by Eqn.~\eqref{eq:select} for each candidate episode
\STATE Select $B$ episode pairs $\{(\mathcal{G}_t^{(1)},\mathcal{G}_t^{(2)})\}_{t=1}^{B}$ from candidates according to $\eta_t$ to form a mini-batch
\FOR{$t=1,\dots,B$}

\STATE Calculate classification loss on support set $\mathcal{L}^{cls}_{t,\mathcal{S}}$ by Eqn.~\eqref{eq:cls_ts} on both $\mathcal{G}_t^{(1)}$ and $\mathcal{G}_t^{(2)}$ 
\STATE Do inner-loop update $\theta^\prime\gets\theta-\beta_{inner} \nabla_{\theta}\mathcal{L}_{t,\mathcal{S}}^{cls}\left(f_{(\theta)}\right)$ on both $\mathcal{G}_t^{(1)}$ and $\mathcal{G}_t^{(2)}$
\STATE Calculate classification loss on query set $\mathcal{L}^{cls}_{t,\mathcal{Q}}$ by Eqn.~\eqref{eq:cls_ts} on both $\mathcal{G}_t^{(1)}$ and $\mathcal{G}_t^{(2)}$

\ENDFOR
\STATE Calculate contrastive loss $\mathcal{L}^{ctr}$ by Eqn.~\eqref{eq:contr}
\STATE Do outer-loop optimization by Eqn.~\eqref{eq:outer}
\STATE Update scheduler $z$ by Eqn.~\eqref{eq:select-opt}
\ENDWHILE
\end{algorithmic}
\end{algorithm}

\subsection{Training and Testing}
\label{sec:process}
In this subsection, we introduce the optimization strategy of relation learning module $f_{(\theta)}$ and subgraph sampling scheduler $z_{(\phi)}$ in training and testing.

\paragraph{Optimization of Relation Learning Module}
Following \cite{DBLP:conf/icml/FinnAL17}, a gradient descent strategy is adopted to obtain a good initialization. Firstly, at the beginning of a mini-batch, subgraph sampling scheduler $z$ is used to sample B episode pairs $\{(\mathcal{G}_t^{(1)},\mathcal{G}_t^{(2)})\}_{t=1}^{B}$ from candidates.

For each episode, in the inner-loop optimization, the loss on the support set defined in Eqn.\eqref{eq:cls_ts} is computed to update the parameters $\theta$ by gradient descent:
\begin{equation}
\label{eq:inner}
    \theta^\prime\gets\theta-\beta_{inner} \nabla_{\theta}\mathcal{L}_{t,\mathcal{S}}^{cls}\left(f_{(\theta)}\right),
\end{equation}
where $\beta_{inner}$ is the learning rate. After updating the parameter, the loss of query set is computed, denoted as ${{\mathcal{L}_{t,\mathcal{Q}}^{cls}}}$. Finally, we do an outer loop to optimize both the classification and contrastive loss with learning rate $\beta_{outer}$ across the mini-batch:
\begin{equation}
\label{eq:outer}
    \theta\gets\theta-\beta_{outer} \nabla_{\theta}\mathcal{L}\left(f_{(\theta^\prime)}\right),
\end{equation}
{where the meta-training loss $\mathcal{L}\left(f_{(\theta^\prime)}\right)$ is computed across the mini-batch:}
\begin{equation}
\label{eq:outer-loss}
        \mathcal{L}\left(f_{(\theta^\prime)}\right)=\frac{1}{2B}\sum_{t=1}^{B}\left({{\mathcal{L}_{t,\mathcal{Q}}^{cls}}}^{(1)}+{{\mathcal{L}_{t,\mathcal{Q}}^{cls}}}^{(2)}\right)+\lambda \mathcal{L}^{ctr},
\end{equation}
where $\lambda$ is hyperparameter and $\mathcal{L}^{ctr}$ is defined by Eqn.\eqref{eq:contr}, and ${{\mathcal{L}_{t,\mathcal{Q}}^{cls}}}^{(1)}$, ${{\mathcal{L}_{t,\mathcal{Q}}^{cls}}}^{(2)}$ are query loss of $\mathcal{G}_t^{(1)}$ and $\mathcal{G}_t^{(2)}$ respectively. The complete procedure is described in Algorithm~\ref{alg:algorithm}.

In testing, for a new property $p_t$ in $\{p_{test}\}$, auxiliary properties are selected from $\{p_{train}\}$ and $f_{(\theta)}$ is finetuned by Eqn.\eqref{eq:inner}.

\paragraph{Optimization of Subgraph Sampling Scheduler}
Since sampling cannot be directly differentiated, similar to \cite{DBLP:conf/nips/YaoWWZMLF21}, we use policy gradient \cite{williams1992simple} to optimize the scheduler $z_{(\phi)}$. {To encourage mining negative samples that are indistinguishable from positives, it is intuitive to take the value of contrastive loss $\mathcal{L}^{ctr}$ as reward $R$}:
\begin{equation}
\label{eq:select-opt}
    \phi \leftarrow \phi+\gamma \nabla_{\phi} \log P
\left(\boldsymbol{\eta}\right)\left({R}-b\right),
\end{equation}
where $\boldsymbol{\eta}=\{\eta_t\}_{t=1}^{B}$ is selection probability of sampled episode in a mini-batch, $\gamma$ is learning rate and $b$ is moving average of reward.

\section{Experiments}
The following research questions guide the remainder of the paper. (\textbf{RQ1}) Can our proposed GS-Meta outperform SOTA baselines? (\textbf{RQ2}) How do auxiliary properties affect the performance? (\textbf{RQ3}) Can the episode reformulation and sampling scheduler improve performance? {(\textbf{RQ4}) How to interpret the scheduler that models the episode relationship?}

\subsection{Experimental Setup}
\label{sec:experiments_set}
We use five common {few-shot molecular property prediction} datasets from the MoleculeNet~\cite{wu2018moleculenet}. Details are in Appendix~\ref{sec:datasets-detail}.

For a comprehensive comparison, we adopt two types of baselines: (1) \textsl{methods with molecular encoder learned from scratch}, including Siamese~\cite{koch2015siamese}, ProtoNet~\cite{DBLP:conf/nips/SnellSZ17}, MAML~\cite{DBLP:conf/icml/FinnAL17}, TPN~\cite{DBLP:conf/iclr/LiuLPKYHY19}, EGNN~\cite{DBLP:conf/cvpr/KimKKY19}, IterRefLSTM~\cite{altae2017low}, and PAR~\cite{DBLP:conf/nips/WangAYD21}; and (2) \textsl{methods which leverage pre-trained molecular encoder}, including Pre-GNN~\cite{DBLP:conf/iclr/HuLGZLPL20}, Meta-MGNN~\cite{DBLP:conf/www/GuoZYHW0C21}, Pre-PAR~\cite{DBLP:conf/nips/WangAYD21} and we denote Pre-GS-Meta as our method equipped with Pre-GNN. More details about these baselines are in Appendix~\ref{sect:baselines}. We run experiments 10 times with different random seeds and report the mean and standard deviations.

\begin{table*}[!ht]
	\small
	\centering
	\setlength\tabcolsep{5pt}

    \scalebox{0.85}{{
	\begin{tabular}{l|cc|cc|cc|cc|cc}
		\hline
		\hline
		\multirow{2}{*}{$\!\!$Method$\!\!$} &         \multicolumn{2}{c|}{Tox21}                 &         \multicolumn{2}{c|}{SIDER}                 &          \multicolumn{2}{c|}{MUV}        &
		\multicolumn{2}{c|}{ToxCast}    &
		\multicolumn{2}{c}{PCBA}  \\
		& \multicolumn{1}{c}{10-shot} & \multicolumn{1}{c|}{1-shot} & \multicolumn{1}{c}{10-shot} & \multicolumn{1}{c|}{1-shot} & \multicolumn{1}{c}{10-shot} & \multicolumn{1}{c|}{1-shot}& \multicolumn{1}{c}{10-shot} & \multicolumn{1}{c|}{1-shot}& \multicolumn{1}{c}{10-shot} & \multicolumn{1}{c}{1-shot}  \\ \hline
		$\!\!$Siamese                 & $\!\!$$80.40_{(0.35)}$$\!\!$                  & $\!\!$$65.00_{(1.58)}$$\!\!$                   & $\!\!$$71.10_{(4.32)}$$\!\!$                   & $\!\!$$51.43_{(3.31)}$$\!\!$                   & $\!\!$$59.96_{(5.13)}$$\!\!$                   & $\!\!$$50.00_{(0.17)}$$\!\!$    &-&- &-&-  
		\\ 
		$\!\!$ProtoNet                & $\!\!$$74.98_{(0.32)}$$\!\!$                   & $\!\!$$65.58_{(1.72)}$$\!\!$                   & $\!\!$$64.54_{(0.89)}$$\!\!$                   & $\!\!$$57.50_{(2.34)}$$\!\!$                   & {$\!\!$$65.88_{(4.11)}$$\!\!$}                    & $\!\!$$58.31_{(3.18)}$$\!\!$   & $\!\!$$68.87_{(0.43)}$$\!\!$  & $\!\!$$58.55_{(0.52)}$$\!\!$       &         
		$\!\!$$64.93_{(1.94)}$$\!\!$  & $\!\!$$55.79_{(1.45)}$$\!\!$                \\
		$\!\!$MAML                    & $\!\!$$80.21_{(0.24)}$$\!\!$                   & $\!\!$$75.74_{(0.48)}$$\!\!$                   & $\!\!$$70.43_{(0.76)}$$\!\!$                   & $\!\!$$67.81_{(1.12)}$$\!\!$                   & $\!\!$$63.90_{(2.28)}$$\!\!$                   & $\!\!$$60.51_{(3.12)}$$\!\!$     & {$\!\!$$68.30_{(0.59)}$$\!\!$}  & {$\!\!$$61.12_{(0.94)}$$\!\!$}          &   
		{$\!\!$$66.22_{(1.31)}$$\!\!$}  & {$\!\!$$62.04_{(1.73)}$$\!\!$}             \\
		$\!\!$TPN                     & $\!\!$$76.05_{(0.24)}$$\!\!$                   & $\!\!$$60.16_{(1.18)}$$\!\!$                   & $\!\!$$67.84_{(0.95)}$$\!\!$                   & $\!\!$$62.90_{(1.38)}$$\!\!$                   & $\!\!$$65.22_{(5.82)}$$\!\!$                   & $\!\!$$50.00_{(0.51)}$$\!\!$     & - & -  &-&-            \\
		$\!\!$EGNN                    & $\!\!$$81.21_{(0.16)}$$\!\!$                  & $\!\!$$79.44_{(0.22)}$$\!\!$                   & $\!\!$$72.87_{(0.73)}$$\!\!$                 & $\!\!$$70.79_{(0.95)}$$\!\!$                   & $\!\!$$65.20_{(2.08)}$$\!\!$                  & $\!\!$$62.18_{(1.76)}$$\!\!$   & $\!\!$$74.02_{(1.11)}$$\!\!$  & $\!\!$$64.17_{(0.89)}$$\!\!$          &
		$\!\!$$69.92_{(1.85)}$$\!\!$  & $\!\!$$62.14_{(1.58)}$$\!\!$                \\ 
		$\!\!$IterRefLSTM$\!\!$             & $\!\!$$81.10_{(0.17)}$$\!\!$                   & $\!\!$$\underline{80.97}_{(0.10)}$$\!\!$                   & $\!\!$$69.63_{(0.31)}$$\!\!$                   & {$\!\!$$71.73_{(0.14)}$$\!\!$}                   & $\!\!$$49.56_{(5.12)}$$\!\!$                   & $\!\!$$48.54_{(3.12)}$$\!\!$     &-&-   &-&-               \\
		$\!\!$PAR                    & {$\!\!$$\underline{82.06}_{(0.12)}$$\!\!$}                   & {$\!\!$$80.46_{(0.13)}$$\!\!$}                   & {$\!\!$$\underline{74.68}_{(0.31)}$$\!\!$}                    & {$\!\!$$\underline{71.87}_{(0.48)}$$\!\!$}                    & {$\!\!$$\underline{66.48}_{(2.12)}$$\!\!$}                    & {$\!\!$$\underline{64.12}_{(1.18)}$$\!\!$}        & {$\!\!$$\underline{74.78}_{(1.53)}$$\!\!$}   & {$\!\!$$\underline{69.45}_{(1.24)}$$\!\!$}    &
		{$\!\!$$\underline{70.05}_{(0.94)}$$\!\!$}    &
		{$\!\!$$\underline{67.77}_{(1.04)}$$\!\!$}   \\
		$\!\!$GS-Meta                    & {$\!\!$$\textbf{85.85}_{(0.26)}$$\!\!$}                   & {$\!\!$$\textbf{84.32}_{(0.89)}$$\!\!$}                   & {$\!\!$$\textbf{83.72}_{(0.54)}$$\!\!$}                    & {$\!\!$$\textbf{82.84}_{(0.67)}$$\!\!$}                    & {$\!\!$$\textbf{67.11}_{(1.95)}$$\!\!$}                    & {$\!\!$$\textbf{64.70}_{(2.88)}$$\!\!$}        & {$\!\!$$\textbf{81.55}_{(0.19)}$$\!\!$}   & {$\!\!$$\textbf{80.03}_{(0.26)}$$\!\!$}   &
		{$\!\!$$\textbf{72.16}_{(0.71)}$$\!\!$}   & {$\!\!$$\textbf{70.03}_{(1.56)}$$\!\!$}     \\
		\hline
		$\!\!$Pre-GNN                  & $\!\!$$82.14_{(0.08)}$$\!\!$                   & $\!\!$$81.68_{(0.09)}$$\!\!$                   & $\!\!$$73.96_{(0.08)}$$\!\!$                   & {$\!\!$$73.24_{(0.12)}$$\!\!$}        & $\!\!$$67.14_{(1.58)}$$\!\!$                   & $\!\!$$64.51_{(1.45)}$$\!\!$      & - & -& -   & -               \\
		$\!\!$Meta-MGNN$\!\!$               & {$\!\!$$82.97_{(0.10)}$$\!\!$}        & {$\!\!$$82.13_{(0.13)}$$\!\!$}        & {$\!\!$$75.43_{(0.21)}$$\!\!$}        & {$\!\!$$73.36_{(0.32)}$$\!\!$}        & {$\!\!$$68.99_{(1.84)}$$\!\!$}        & {$\!\!$$65.54_{(2.13)}$$\!\!$}      &-&- &-&-   \\
		$\!\!$Pre-PAR                & {$\!\!$$\underline{84.93}_{(0.11)}$$\!\!$}            & {$\!\!$$\underline{83.01}_{(0.09)}$$\!\!$}          & {$\!\!$$\underline{78.08}_{(0.16)}$$\!\!$}            & {$\!\!$$\underline{74.46}_{(0.29)}$$\!\!$}          & {$\!\!$$\underline{69.96}_{(1.37)}$$\!\!$}         & {$\!\!$$\underline{66.94}_{(1.12)}$$\!\!$}     & {$\!\!$$\underline{79.41}_{(0.08)}$$\!\!$}   & {$\!\!$$\underline{76.58}_{(0.15)}$$\!\!$}   & {$\!\!$$\underline{73.71}_{(0.61)}$$\!\!$}   & {$\!\!$$\underline{72.49}_{(0.61)}$$\!\!$}       \\
		$\!\!$Pre-GS-Meta                & {$\!\!$$\textbf{86.91}_{(0.41)}$$\!\!$}            & {$\!\!$$\textbf{86.46}_{(0.55)}$$\!\!$}          & {$\!\!$$\textbf{85.08}_{(0.54)}$$\!\!$}            & {$\!\!$$\textbf{84.45}_{(0.26)}$$\!\!$}           & {$\!\!$$\textbf{70.18}_{(1.25)}$$\!\!$}          & {$\!\!$$\textbf{67.15}_{(2.04)}$$\!\!$}     & {$\!\!$$\textbf{83.81}_{(0.16)}$$\!\!$}   & {$\!\!$$\textbf{81.57}_{(0.18)}$$\!\!$}   &
		{$\!\!$$\textbf{79.40}_{(0.43)}$$\!\!$}   & {$\!\!$$\textbf{78.16}_{(0.47)}$$\!\!$}\\
		\hline
		\hline
	\end{tabular}
	}}
\caption{ROC-AUC scores on benchmark datasets, compared with methods learned from scratch (first group) and methods that leverage pre-trained molecular encoder (second group). The best is marked with \textbf{boldface} and the second best is with \underline{underline}.
	}
\label{tab:exp-fsl}
\end{table*}

\subsection{Main Results (RQ1)}
\label{sec:main_results}
The overall performance is shown in Table~\ref{tab:exp-fsl}. The experimental results show that our GS-Meta and Pre-GS-Meta outperform all baselines consistently with 
{6.93}\% and {5.71}\% average relative improvement respectively. {Moreover, those relation-graph-based methods (\textit{i.e.,} TPN, EGNN, PAR and GS-Meta) mostly perform better than traditional methods.} This indicates that it is effective to exploit relations between molecules in a graph and it is more effective to exploit both molecules and properties together as our method performs best. Further, the improvement varies across different datasets. For example, GS-Meta gains {13.68}\% improvement on SIDER but {0.93}\% improvement on MUV. We argue this is due to the fact that there are more auxiliary properties available on SIDER, and also that there are no missing labels in SIDER compared to 84.2\% missing labels on MUV. Fewer auxiliary properties and the presence of missing labels prevent utilizing information from auxiliary properties. This phenomenon is further investigated in Section~\ref{sec:auxi-ana}.

\begin{figure}[!t]
\centering
\includegraphics[width=0.95\columnwidth]{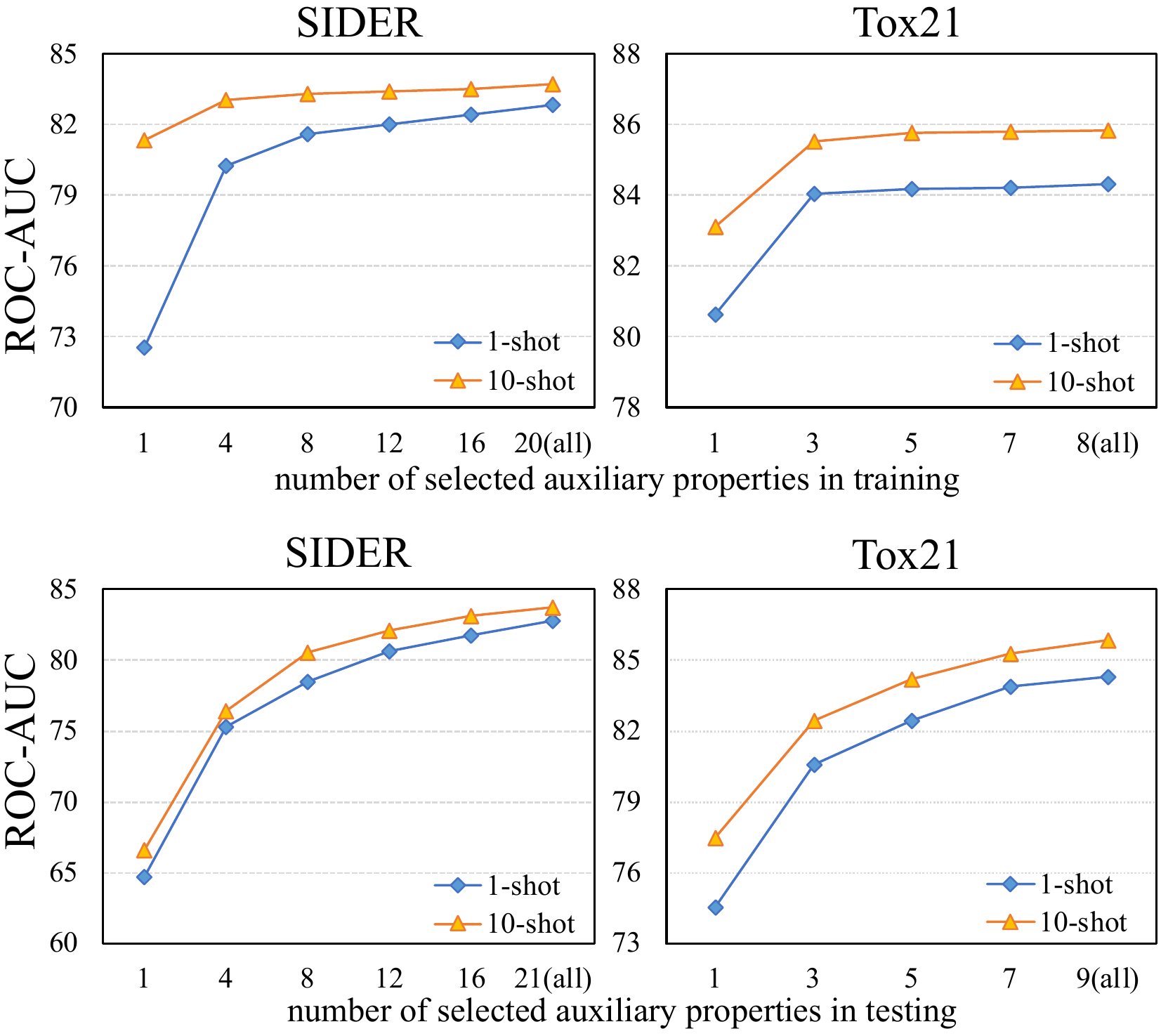} 
\caption{Performance of different numbers of selected auxiliary properties in training (top) and testing (bottom).}
\label{fig:auxi}
\end{figure}

\subsection{Analysis of Auxiliary Property (RQ2)}
\label{sec:auxi-ana}
\paragraph{Effect of the Number of Auxiliary Properties}
We explore the effect of auxiliary properties by varying the number of 
sampled auxiliary properties. Since auxiliary property sampling occurs during both training and testing, here we consider two scenarios: 1) \textbf{effect of the number of sampled auxiliary properties during training}: keep the number of sampled auxiliary properties during testing constant and change the number during training; 2) \textbf{effect of the number of sampled auxiliary properties during testing}: keep the number of sampled auxiliary properties during training constant and change the number during testing. 

As shown in Figure~\ref{fig:auxi}, the performance improves as the number of auxiliary properties increases in both training and testing, confirming our motivation that known properties of molecules help predict new properties. This is because more auxiliary properties contain more information at each training and inference step. In addition, reducing the number of auxiliary properties in training has less impact on performance than in testing,
which suggests that when faced with a large number of auxiliary properties, sampling some of them during training can be an effective way to train the model and does not lead to significant performance degradation.

\begin{figure}[!t]
\centering
\includegraphics[width=\columnwidth]{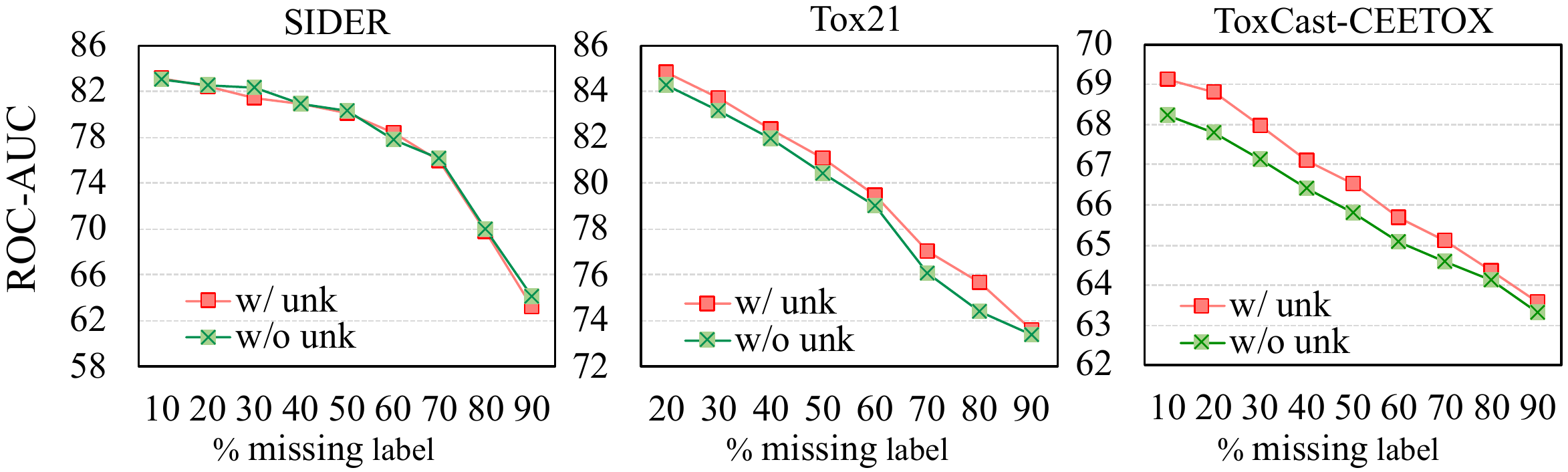} 
\caption{Performance with different missing label ratios in SIDER, Tox21 and ToxCast-CEETOX under the 10-shot scenario.}
\label{fig:missing-10shot}
\end{figure}

\paragraph{Effect of the Ratio of Missing Labels}
We further investigate the effect of missing labels on auxiliary properties by randomly masking labels in the training set, which means the model has less training data and fewer auxiliary properties possible for testing. Two different approaches are compared: 1) \textbf{w/ \textit{unk}}: completing missing label with a special edge \textit{unk}; 2) \textbf{w/o \textit{unk}}: no process of the missing label.

Figure~\ref{fig:missing-10shot} shows the results under the 10-shot scenario. Since the proportion of missing labels in the training set on Tox21 itself is higher than 10\%, we start masking from 20\%.
The performance gradually decreases as the percentage of missing labels increases. In addition, \textit{unk} edges have no noticeable effect on SIDER,
but improve performance on Tox21 and ToxCast-CEETOX by 0.81\% and 0.96\% respectively.
This can be due to imbalanced labeling: SIDER is more balanced than Tox21 and ToxCast-CEETOX. Moreover, the results show that when masking 70\%, 40\% and 40\% training labels on SIDER, Tox21 and ToxCast-CEETOX respectively, our method still achieves a comparable performance against SOTA. It proves that our method has {robust} and promising performance despite the missing training data. Results under the 1-shot scenario is in Figure~\ref{fig:missing-1shot} in Appendix.

\subsection{Analysis of Episode Reformulation and Sampling Scheduler (RQ3)}

\paragraph{Ablation Study}
We conduct ablations to study the effectiveness of episode reformulation and sampling scheduler. For episode reformulation, the following variants are analyzed: 1)\textbf{w/o m2m}: remove \textit{mol2mol} edges in MPG; 2)\textbf{w/o E}: remove edge type in MPG, \textit{i.e.}, do message passing in Eqn.\eqref{eq:message} without $\boldsymbol{h}_{i,j}$. For the sampling scheduler, the following variants are analyzed: 1)\textbf{w/o S}: randomly select episode without a sampling scheduler; 2)\textbf{w/o CL}: optimize model parameters in Eqn.~\eqref{eq:outer} without contrastive loss $\mathcal{L}^{ctr}$; 3)\textbf{w/o S, w/o CL}: do not use scheduler and contrastive loss.

As in Table~\ref{tab:ablation}, GS-Meta outperforms all the variants, indicating that our proposed subgraph reformulation and scheduling strategy are effective. {Results on SIDER are in Table~\ref{tab:ablation-sider} in Appendix}. Removing the type of edges causes a significant performance drop, which validates that the information of labels can be fused into the graph by using them as edge types. And removing both contrastive loss and the scheduler degrades the model performance, indicating that our design of contrastive loss and the scheduler is reasonably effective.

\begin{table}[!t]
\centering


\setlength\tabcolsep{12pt}
\scalebox{0.8}{
{
\begin{tabular}{l|cc}
\hline

Method & 1-shot & 10-shot \\
\hline
{GS-Meta} & \textbf{84.32} & \textbf{85.85} \\
w/o m2m & 83.91($\downarrow$0.41) & 84.80($\downarrow$1.05) \\
w/o E & 72.54($\downarrow$11.78) & 75.93($\downarrow$9.92) \\
w/o S & 83.14($\downarrow$1.18) & 84.51($\downarrow$1.34) \\
w/o CL & 83.30($\downarrow$1.02) & 84.66($\downarrow$1.19) \\
w/o S, w/o CL & 82.64($\downarrow$1.68) & 84.32($\downarrow$1.53) \\
\hline
\end{tabular}
}
}
\caption{Ablation study on Tox21 }
\label{tab:ablation}
\end{table}

\begin{table}[!t]
\centering

\scalebox{0.8}{
{
\begin{tabular}{l|cc}
\hline

Method & 1-shot & 10-shot \\
\hline
PAR & 70.41(0.83s) & 74.65(1.13s) \\
PAR (w/ ATS) & 69.94(2.42s) & 75.10(2.90s) \\
\hline
GS-Meta & \textbf{85.94}(3.11s) & \textbf{87.67}(3.59s) \\
GS-Meta (w/ ATS) & 84.73(6.80s) & 86.88(7.11s)\\
GS-Meta (w/o S) &84.72(2.20s)&86.81(2.60s)\\
\hline
\end{tabular}
}
}
\caption{Performance and time cost on ToxCast-BSK}
\label{tab:scheduler}

\end{table}

\paragraph{Comparing with Other Scheduler}
To further explore the effects of the designed sampling scheduler, we compare PAR and our GS-Meta with other scheduler. Here we adopt ATS~\cite{DBLP:conf/nips/YaoWWZMLF21}, which is a SOTA task scheduler for meta-learning proposed recently. ATS takes gradient and loss as input to characterize the difficulty of candidate tasks. For a comprehensive evaluation, we compare both the performance and time cost of a mini-batch during training. 

The results on ToxCast-BSK are in Table~\ref{tab:scheduler}, where (w/ ATS) indicates using ATS as the sampling scheduler and (w/o S) indicates randomly selecting without the scheduler. Note that the original PAR itself does not have a scheduler. We reach two conclusions.
First, ATS does not improve performance significantly (\textit{PAR(w/ ATS)} {vs} \textit{PAR}, \textit{GS-Meta(w/ ATS)} {vs} \textit{GS-Meta(w/o S)}).
This suggests that ATS is not applicable in our scenario of few-shot molecular property prediction and the SOTA methods. And our scheduler is able to improve the model performance (\textit{GS-Meta} {vs} \textit{GS-Meta(w/o S)}). Secondly, ATS is more time-consuming. ATS is around three times slower compared with random sampling (\textit{PAR(w/ ATS)} {vs} \textit{PAR}, \textit{GS-Meta(w/ ATS)} {vs} \textit{GS-Meta(w/o S)}), but our scheduler is faster (\textit{GS-Meta} {vs} \textit{GS-Meta(w/o S)}).
This is because ATS needs to compute the loss and gradient backpropagation for each candidate and sample a validation set to get the reward for optimization. While our approach is to reformulate an episode as a subgraph and get the representation of episode by directly encoding it. And we use a contrastive loss to uniformly model the relationship between episodes and as an optimization reward for the scheduler. Results on ToxCast-APR are in Table~\ref{tab:scheduler-APR} in Appendix.

\subsection{Interpretation of the Scheduler (RQ4)}
\begin{figure}[!t]
\centering
\includegraphics[width=\columnwidth]{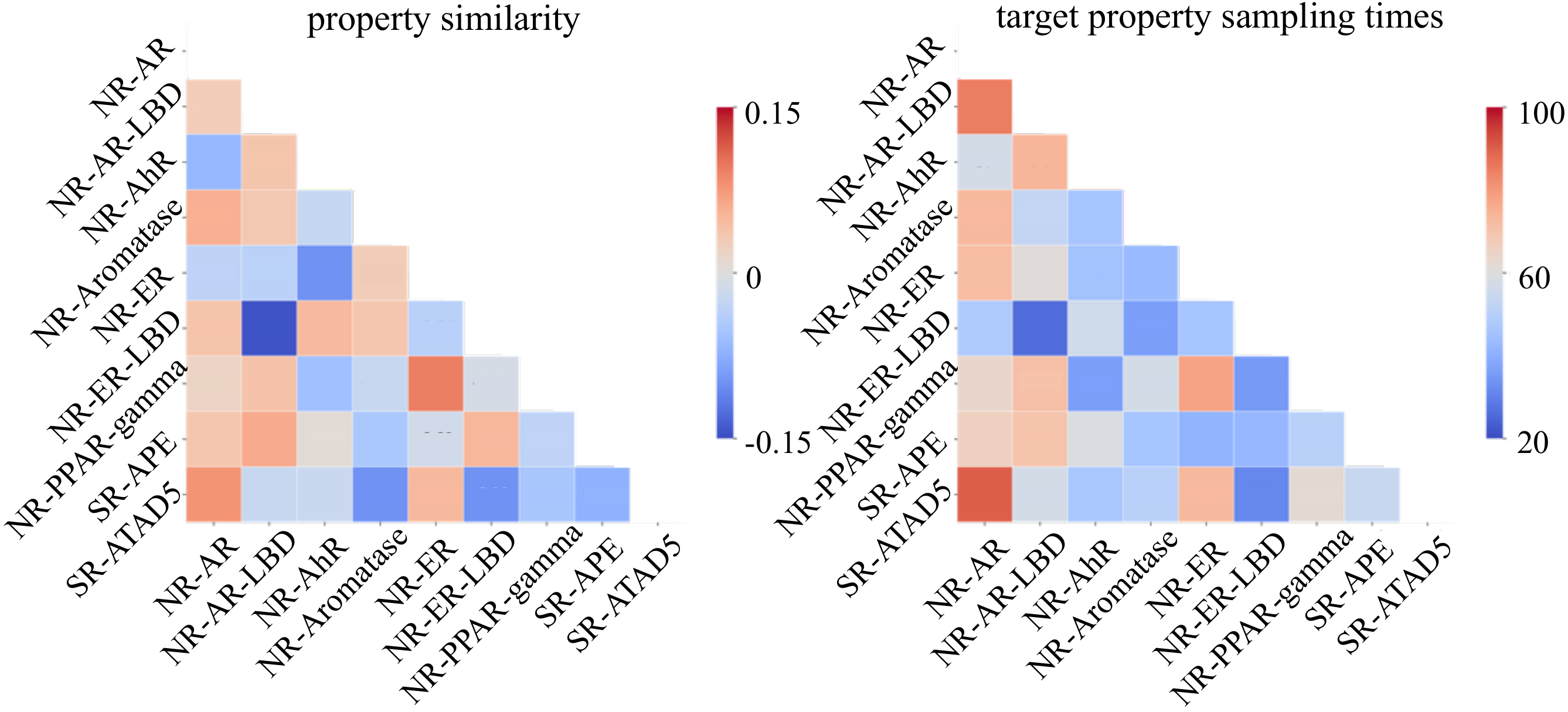} 
\caption{Heatmap of property similarity and statistics of target property sampling times on Tox21.}
\label{fig:case}
\end{figure}

To understand the sampling scheduler, we visualize the sampling result of 9 properties in the training set on Tox21 in Figure~\ref{fig:case}. The cosine similarity of property embeddings on Tox21 is also visualized. In the sampling result heatmap, the value in row \textit{i} column \textit{j} indicates the number of times that property \textit{i} and property \textit{j} are sampled in the same mini-batch at the same time. And we run the scheduler 200 times to statistically count the results.

We observe that {more} similar properties are sampled into the same mini-batch more frequently, \textit{e.g.}, property {SR-ATAD5} with property {NR-AR} and property {NR-ER} with property {NR-PPAR-gamma}. And dissimilar properties are sampled less frequently in one mini-batch, \textit{e.g.}, property {NR-ER-LBD} with property {NR-AR-LBD} and property {SR-ATAD5} with property {NR-ER-LBD}. This indicates that our scheduler prefers to put similar properties in one mini-batch, {analogous to mining hard negative samples.}
{Nevertheless,} the sampling result is not exactly consistent with the property embedding similarity, because the scheduler's input is the final pooled embedding of the subgraph, not only the embedding of the target property.

\section{Conclusion}
We propose a Graph Sampling-based Meta-learning framework to effectively leverage other available properties to predict new properties with a few labeled samples in molecular property prediction. Specifically, we construct a Molecule-Property relation Graph and reformulate an episode in meta-learning as a subgraph of MPG. Further, we  consider subgraph consistency and discrimination and schedule the subgraph sampling via a contrastive loss. Empirical results show GS-Meta outperforms previous methods consistently. 

This work only considers 2-way classification tasks and does not involve regression tasks. This is because the commonly-used benchmarks for few-shot molecular property prediction are 2-way tasks and few are eligible regression datasets. Our method can generalize to multi-class and regression cases, by assigning different edge attribute values. We put this in future works.

\newpage
\section*{Acknowledgments}
This work was supported by the National Natural Science Foundation of China (NSFCU19B2027, NSFC91846204), joint project DH-2022ZY0012 from Donghai Lab,
and CAAI-Huawei MindSpore Open Fund (CAAIXSJLJJ-2022-052A). We want to express gratitude to the anonymous reviewers for their hard work and kind comments and Hangzhou AI Computing Center for their technical support.

\bibliographystyle{named}
\bibliography{main}

\clearpage
\input{appendix}

\end{document}

%% file: appendix.tex
\appendix
\section*{Appendix}
\section{Details of GS-Meta}

\subsection{Notations}
\label{sec:notation}
For ease of understanding, we summarize notations and descriptions in Table~\ref{tab:symbol}.

\subsection{Reasons to Sampling Subgraphs as Episodes}
The size of the constructed MPG can be very large (\textit{e.g.}, containing more than 400,000 molecules in PCBA dataset), and it is not possible to compute direcly on the whole graph during training. Because the molecular encoder $f_{mol}$ needs to encode each molecule to get the initial embedding of molecular nodes in the graph. Encoding molecules using molecular encoder and encoding MPG using GNN, and then doing gradient back propagation will lead to an Out-of-Memory error.

\subsection{GNN Layer}
\label{sec:msg-detail}
Eqn.~\eqref{eq:message} follows a message passing paradigm, in which embedding is updated by aggregating the information passed from neighboring nodes and edges. At $l$-th iteration, firstly the aggregated neighborhood embedding $\boldsymbol{h}_{\mathcal{N}_{(i)}}^l$ is obtained:
\begin{equation}
\boldsymbol{h}^{l}_{\mathcal{N}_{(i)}}\gets\frac{1}{|\mathcal{N}_{(i)}|}\sum_{j\in\mathcal{N}_{(i)}}\left(\left(\boldsymbol{h}_j^{l-1}+\boldsymbol{h}_{i,j}^{l-1}\right) \times w_{i,j}^{l} \right)
\end{equation}
where $\mathcal{N}_{(i)}$ is neighbors of node $i$, $\boldsymbol{h}_{i,j}^{l-1}$ and $w_{i,j}^l$ is embedding and weight of edge. And then we do combination to process information received from neighborhood and node's previous layer embedding $\boldsymbol{h}_i^{l-1}$:
\begin{equation}
    \boldsymbol{h}^l_i \gets \mathrm{LeakyReLU}\left(\mathbf{W}^{(l)}_{\mathrm{msg}}\boldsymbol{h}^{l}_{\mathcal{N}_{(i)}} + \mathbf{W}^{(l)}_{\mathrm{root}}\boldsymbol{h}_i^{l-1}\right),
\end{equation}
where $\mathbf{W}^{(l)}_{\mathrm{msg}}$ and $\mathbf{W}^{(l)}_{\mathrm{root}}$ are trainable parameters included in parameters $\theta$ of relation learning module $f_{(\theta)}$.

\begin{table}[h]
\small

\centering
\scalebox{0.88}{
{
\begin{tabular}{c|ccccc}
\hline
Dataset               & Tox21 & SIDER & MUV & ToxCast &PCBA \\ \hline
\#Compound           &  7831   &   1427  & 93127   &  8575 &437929   \\
\#Property      & 12   & 27   &  17 & 617 & 128 \\
\#Train Property &  9  &   21  & 12  & 451 &118     \\
\#Test Property &  3   &6  &  5 &158  &10  \\ \hline
\%Label \textit{active} &6.24&56.76&0.31&12.60&0.84 \\
\%Label \textit{inactive} &76.71&43.24&15.76&72.43&59.84 \\
\%Missing Label &17.05&0&84.21&14.97&39.32 \\
\hline 
\end{tabular}
}
}
\caption{Dataset statistics}
\label{tab:dataset}
\end{table}

\section{Details of Datasets}
\label{sec:datasets-detail}
\subsection{Datasets Description}
We conduct experiments on five widely used few-shot molecular property prediction datasets(Table~\ref{tab:dataset}) in MoleculeNet benchmark~\cite{wu2018moleculenet}:
\begin{itemize}
    \item Tox21\footnote{\url{https://tripod.nih.gov/tox21/challenge/}} is a public database on compound toxicity, which has been used in the 2014 Tox21 Data Challenge.
    \item SIDER~\cite{DBLP:journals/nar/KuhnLJB16} contains marketed drugs and adverse drug reactions(ADR), which are grouped into 27 system organ classes.
    \item MUV~\cite{rohrer2009maximum} is selected by applying a redefined nearest neighbor analysis for validation of virtual screening techniques.
    \item ToxCast~\cite{richard2016toxcast} provides toxicology data for a large quantities of compounds based on in vitro high-throughput screening.
    \item PCBA~\cite{wang2012pubchem} consists of small molecule bioactivities generated by high-throughput screening.
\end{itemize}

\begin{table}[!t]
\centering
\small
\caption{Notations and Description}
\label{tab:symbol}
\setlength{\tabcolsep}{3pt}
\vspace{-3mm}
\begin{tabular}{|l|l|}
\hline
\textbf{Notation} & \textbf{Description} \\
\hline
$\mathcal{D}_{test}$ &  Training data set \\ \hline
$\mathcal{D}_{test}$ &  Testing data set \\ \hline
$\{\mathcal{T}_{train}\}$ & Training task set \\ \hline
$\{\mathcal{T}_{test}\}$ & Testing task set \\ \hline
$\mathcal{S}_{t},\mathcal{Q}_{t}$ & Support and query set of task $\mathcal{T}_t$ \\ \hline
$E_t$ &  One episode of task $\mathcal{T}_t$ \\ \hline
$\mathcal{G}$ &  Sample-Property relation Graph \\ \hline
$\mathcal{V}$,$\mathcal{E}$ & Node and edge set of $\mathcal{G}$ \\ \hline
$\mathcal{V}_T$        & Property node set \\ \hline
$\mathcal{V}_M$        & Molecule node set \\ \hline
$p_t$        &  The $t$-th property node in $\mathcal{G}$ \\ \hline
$x_i$          &  The $i$-th molecule node in $\mathcal{G}$ \\ \hline
$\{p_{train}\}$        & property nodes corresponding to $\{\mathcal{T}_{train}\}$ \\ \hline
$\{p_{test}\}$        & property nodes corresponding to $\{\mathcal{T}_{test}\}$ \\ \hline
$\boldsymbol{p_t}$        &  Initial embedding of $t$th property node\\ \hline
$\boldsymbol{x_i}$        &  Initial embedding of $i$th molecule node\\ \hline
$e_{i,j}$        & Edge between node $i$ and node $j$ \\ \hline
$\mathcal{G}_S^t$        &  Subgraph in $\mathcal{G}$ analogous to $\mathcal{S}_t$ \\ \hline
$\mathcal{G}_Q^t$        &  Subgraph in $\mathcal{G}$ analogous to $\mathcal{Q}_t$ \\ \hline
$\mathcal{G}_A^t$        &  Auxiliary subgraph in $\mathcal{G}$ \\ \hline
$\mathcal{G}_t$        &  Subgraph in $\mathcal{G}$ analogous to $E_t$ \\ \hline
$\boldsymbol{h}_i^l$ &  Embedding of node $i$ at $l$-th layer \\ \hline
$\boldsymbol{h}_{i,j}^l$ &  Embedding of edge $(i,j)$ at $l$-th layer \\ \hline
$\alpha_{i,j}^l$ &  Estimated weight of molecule edge $(i,j)$ at $l$-th layer \\ \hline
$w_{i,j}^l$ &  Edge weight of $(i,j)$ at $l$-th layer \\ \hline
$\hat{y}_{i,t}$ &  Predicted label of molecule $i$ on property $t$ \\\hline
$f_{(\theta)}$ & relation learning module with parameter $\theta$ \\\hline
$z_{(\phi)}$ & the sampling scheduler with parameter $\phi$\\\hline
$\boldsymbol{g}_t$ & subgraph embedding of $\mathcal{G}_t$\\\hline
$\eta_t$ & selection propability\\\hline
$\mathcal{L}^{ctr}$ & Contrastive loss in a minibatch \\\hline
${\mathcal{L}_{t,\mathcal{Q}}^{cls}}$ & Classification loss of query set on episode $\mathcal{G}_t$ \\\hline
${\mathcal{L}_{t,\mathcal{S}}^{cls}}$ & Classification loss of support set on episode $\mathcal{G}_t$ \\\hline
\end{tabular}
\end{table}

\begin{table*}[h]

    \small
    \centering
    \setlength\tabcolsep{3.5pt}
    \vspace{-0.5em}
\scalebox{0.75}{
{
    \begin{tabular}{c|c|c|c|c|c|c|c}
    \hline
    Assay Provider & \#Compound & \#Property & \#Train Property & \#Test Property &\%Label \textit{active} &\%Label \textit{inactive} &\%Missing Label \\
    \hline
    APR & 1039 &43&  33 & 10 &10.30&61.61&28.09 \\
    ATG & 3423 &146& 106 & 40 &5.92&93.92&0.16\\
    BSK & 1445 &115& 84 & 31 & 17.71&82.29&0 \\
    CEETOX & 508 &14& 10 & 4 & 22.26&76.38&1.36 \\
    CLD & 305 &19& 14 & 5 & 30.72&68.30&0.98\\
    NVS & 2130&139 & 100 & 39 & 3.21&4.52&92.27 \\
    OT & 1782&15 & 11 & 4 & 9.78&87.78&2.44\\
    TOX21 & 8241&100 & 80 & 20 & 5.39 & 86.26 & 8.35 \\
    Tanguay & 1039&18 & 13 & 5 & 8.05&90.84&1.11 \\
    \hline
    \end{tabular}
}}
\vspace{-1em}
    \caption{Details of sub-datasets of ToxCast.}
    \label{table:toxcast-detail}

\end{table*}

\begin{table*}[!t]
\small
  \centering
  \vspace{-0.5em}
\scalebox{0.85}{
{
    \begin{tabular}{c|c|c|c}
    \hline
    hyper-parameter & Description & Range & Selected \\
    \hline
    $d$ & dimension of molecule and property embedding. & 300 & 300\\
    $L$ & number of GNN layer. & 1$\sim$3 & 2\\
    $\beta_{inner}$ & learning rate in inner-loop. & 0.01$\sim$0.5 & 0.05\\
    $\beta_{outer}$ & learning rate in outer-loop. & 0.001 & 0.001\\
    $\gamma$ & learning rate of subgraph sampling scheduler. & 0.0001$\sim$0.001 & 0.0005\\
    $\tau$ & temperature in Eqn.~\eqref{eq:contr} . & 0.05$\sim$0.5 & 0.08\\
    $\lambda$ & regularization of contrastive loss in Eqn.~\eqref{eq:contr}. & 0.01$\sim$0.5 & 0.05\\
    \hline
    \end{tabular}%
}}
\vspace{-1em}

  \caption{The hyper-parameters.}
  \label{tab:hyper-param}
\end{table*}

\begin{table*}[!ht]

	\small
	\centering
	\setlength\tabcolsep{9pt}

\scalebox{0.85}{
{
	\begin{tabular}{l|c|c|c|c|c|c|c|c|c}
		\hline
		\hline
		\multirow{1}{*}{Method} &         \multicolumn{1}{c|}{APR}                 &         \multicolumn{1}{c|}{ATG}                 &          \multicolumn{1}{c|}{BSK}        &
		\multicolumn{1}{c|}{CEETOX}    &
		\multicolumn{1}{c|}{CLD} &  \multicolumn{1}{c|}{NVS} &  \multicolumn{1}{c|}{OT} & \multicolumn{1}{c|}{TOX21} &   \multicolumn{1}{c}{Tanguay} 
		\\ \hline
		MAML                     & $\!\!$$64.59$$\!\!$                   & $\!\!$$55.45$$\!\!$                   & $\!\!$$60.36$$\!\!$                   & $\!\!$$61.02$$\!\!$                   & $\!\!$$66.22$$\!\!$                   & $\!\!$$59.84$$\!\!$      & $\!\!$$62.15$$\!\!$     & 
		$\!\!$$59.52$$\!\!$          &   
		$\!\!$$60.92$$\!\!$              \\
        ProtoNet                   & $\!\!$$57.08$$\!\!$                   & $\!\!$$54.92$$\!\!$                   & $\!\!$$53.92$$\!\!$                   & $\!\!$$60.25$$\!\!$                   & $\!\!$$66.25$$\!\!$                   & $\!\!$$54.87$$\!\!$      & $\!\!$$63.11$$\!\!$     & 
		$\!\!$$58.27$$\!\!$          &   
		$\!\!$$58.32$$\!\!$              \\
		EGNN                      & $\!\!$$67.06$$\!\!$                  & $\!\!$$57.28$$\!\!$                   & $\!\!$$60.82$$\!\!$                 & $\!\!$$60.10$$\!\!$                   &
		$\!\!$$71.53$$\!\!$          & $\!\!$$56.56$$\!\!$                  & $\!\!$$66.08$$\!\!$   &     $\!\!$$63.32$$\!\!$       & $\!\!$$74.80$$\!\!$                      \\
		PAR                    & {$\!\!$$\underline{74.24}$$\!\!$}          & {$\!\!$$\underline{63.48}$$\!\!$}                   & {$\!\!$$\underline{70.41}$$\!\!$}   & {$\!\!$$\underline{61.44}$$\!\!$}         & {$\!\!$$\underline{75.76}$$\!\!$}         &
		{$\!\!$$\underline{67.56}$$\!\!$}  &{$\!\!$$\underline{65.72}$$\!\!$}        &
		{$\!\!$$\underline{68.94}$$\!\!$}& {$\!\!$$\underline{77.54}$$\!\!$}     \\
		GS-Meta                    & {$\!\!$$\textbf{87.90}$$\!\!$}                   & {$\!\!$$\textbf{79.62}$$\!\!$}                   & {$\!\!$$\textbf{85.94}$$\!\!$}                    & {$\!\!$$\textbf{67.49}$$\!\!$}                    & {$\!\!$$\textbf{78.16}$$\!\!$}                    & {$\!\!$$\textbf{71.04}$$\!\!$}        & {$\!\!$$\textbf{72.36}$$\!\!$}   & {$\!\!$$\textbf{87.84}$$\!\!$}   &
		{$\!\!$$\textbf{89.97}$$\!\!$}    \\
		\hline
		Pre-PAR                & {$\!\!$$\underline{84.69}$$\!\!$}            & {$\!\!$$\underline{70.38}$$\!\!$}          & {$\!\!$$\underline{79.89}$$\!\!$}            & {$\!\!$$\underline{66.57}$$\!\!$}          & {$\!\!$$\underline{77.83}$$\!\!$}         & {$\!\!$$\underline{72.51}$$\!\!$}     & {$\!\!$$\underline{70.41}$$\!\!$}   & {$\!\!$$\underline{80.33}$$\!\!$}   & {$\!\!$$\underline{86.64}$$\!\!$}      \\
		Pre-GS-Meta                & {$\!\!$$\textbf{89.49}$$\!\!$}            & {$\!\!$$\textbf{81.69}$$\!\!$}          & {$\!\!$$\textbf{87.28}$$\!\!$}            & {$\!\!$$\textbf{68.55}$$\!\!$}           & {$\!\!$$\textbf{78.69}$$\!\!$}          & {$\!\!$$\textbf{74.36}$$\!\!$}     & {$\!\!$$\textbf{73.56}$$\!\!$}   & {$\!\!$$\textbf{89.46}$$\!\!$}   &
		{$\!\!$$\textbf{91.10}$$\!\!$}  \\
		\hline
		\hline
	\end{tabular}
}}
\vspace{-1em}
  \caption{Detailed performance on each sub-dataset of ToxCast in 1-shot scenario.}
		\label{tab:toxcast-sub1shot}

\end{table*}

\begin{table*}[!ht]

	\small
	\centering
	\setlength\tabcolsep{9pt}

\scalebox{0.85}{
{
	\begin{tabular}{l|c|c|c|c|c|c|c|c|c}
		\hline
		\hline
		\multirow{1}{*}{Method} &         \multicolumn{1}{c|}{APR}                 &         \multicolumn{1}{c|}{ATG}                 &          \multicolumn{1}{c|}{BSK}        &
		\multicolumn{1}{c|}{CEETOX}    &
		\multicolumn{1}{c|}{CLD} &  \multicolumn{1}{c|}{NVS} &  \multicolumn{1}{c|}{OT} & \multicolumn{1}{c|}{TOX21} &   \multicolumn{1}{c}{Tanguay} 
		\\ \hline
		MAML                     & $\!\!$$72.66$$\!\!$                   & $\!\!$$62.09$$\!\!$                   & $\!\!$$66.42$$\!\!$                   & $\!\!$$64.08$$\!\!$                   & $\!\!$$74.57$$\!\!$                   & $\!\!$$66.56$$\!\!$      & $\!\!$$64.07$$\!\!$     & 
		$\!\!$$68.04$$\!\!$          &   
		$\!\!$$77.12$$\!\!$              \\
        ProtoNet                   & $\!\!$$73.58$$\!\!$                   & $\!\!$$59.26$$\!\!$                   & $\!\!$$70.15$$\!\!$                   & $\!\!$$66.12$$\!\!$                   & $\!\!$$78.12$$\!\!$                   & $\!\!$$65.85$$\!\!$      & $\!\!$$64.90$$\!\!$     & 
		$\!\!$$68.26$$\!\!$          &   
		$\!\!$$73.61$$\!\!$              \\
		EGNN                      & $\!\!$$80.33$$\!\!$                  & $\!\!$$66.17$$\!\!$                   & $\!\!$$73.43$$\!\!$                 & $\!\!$$66.51$$\!\!$                   &
		$\!\!$$\underline{78.85}$$\!\!$          & $\!\!$$\underline{71.05}$$\!\!$                  & $\!\!$$68.21$$\!\!$   &     $\!\!$$76.40$$\!\!$       & $\!\!$$\underline{85.23}$$\!\!$                      \\
		PAR                    & {$\!\!$$\underline{82.74}$$\!\!$}          & {$\!\!$$\underline{68.86}$$\!\!$}                   & {$\!\!$$\underline{74.65}$$\!\!$}   & {$\!\!$$\underline{67.76}$$\!\!$}         & {$\!\!$$78.33$$\!\!$}         &
		{$\!\!$$70.79$$\!\!$}  &{$\!\!$$\underline{69.12}$$\!\!$}        &
		{$\!\!$$\underline{77.34}$$\!\!$}& {$\!\!$$83.39$$\!\!$}     \\
		GS-Meta                    & {$\!\!$$\textbf{88.95}$$\!\!$}            & {$\!\!$$\textbf{80.44}$$\!\!$}          & {$\!\!$$\textbf{87.67}$$\!\!$}            & {$\!\!$$\textbf{69.50}$$\!\!$}           & {$\!\!$$\textbf{79.95}$$\!\!$}          & {$\!\!$$\textbf{74.77}$$\!\!$}     & {$\!\!$$\textbf{73.46}$$\!\!$}   & {$\!\!$$\textbf{88.78}$$\!\!$}   &
		{$\!\!$$\textbf{90.48}$$\!\!$}    \\
		\hline
		Pre-PAR                & {$\!\!$$\underline{86.09}$$\!\!$}            & {$\!\!$$\underline{72.72}$$\!\!$}          & {$\!\!$$\underline{82.45}$$\!\!$}            & {$\!\!$$\underline{72.12}$$\!\!$}          & {$\!\!$$\underline{83.43}$$\!\!$}         & {$\!\!$$\underline{74.94}$$\!\!$}     & {$\!\!$$\underline{71.96}$$\!\!$}   & {$\!\!$$\underline{82.81}$$\!\!$}   & {$\!\!$$\underline{88.20}$$\!\!$}      \\
		Pre-GS-Meta                & {$\!\!$$\textbf{90.15}$$\!\!$}            & {$\!\!$$\textbf{82.54}$$\!\!$}          & {$\!\!$$\textbf{88.21}$$\!\!$}            & {$\!\!$$\textbf{74.19}$$\!\!$}           & {$\!\!$$\textbf{86.34}$$\!\!$}          & {$\!\!$$\textbf{76.29}$$\!\!$}     & {$\!\!$$\textbf{74.47}$$\!\!$}   & {$\!\!$$\textbf{90.63}$$\!\!$}   &
		{$\!\!$$\textbf{91.47}$$\!\!$}  \\
		\hline
		\hline
	\end{tabular}
}}

\vspace{-1em}
		\caption{Detailed performance on each sub-dataset of ToxCast in 10-shot scenario.}
		\label{tab:toxcast-sub10shot}
\end{table*}

\subsection{Datasets Splitting}
We adopt public splits provided by ~\cite{altae2017low} on Tox21, SIDER and MUV. For PCBA, we choose the first 5 and last 5 properties as meta-testing and the rest of properties as meta-training. For ToxCast, since the dataset is sparse overall, but the properties can be grouped according to assay providers, and the grouping gives denser results. We first group the dataset by assay providers to obtain a number of sub-datasets, and after discarding sub-datasets with few properties, each sub-dataset is divided into meta-training and meta-testing and the average of the performance of all sub-datasets is finnaly reported. Details of each sub-dataset are shown in Table~\ref{table:toxcast-detail} and detailed results of each sub-dataset are in Table~\ref{tab:toxcast-sub1shot} and Table~\ref{tab:toxcast-sub10shot}.

\section{Details of Implementation}
All experiments are conducted on a Ubuntu Server with one 32 GB NVIDIA Tesla V100 GPU.
\subsection{Baselines}
\label{sect:baselines}
We adopt two types of baselines, and details of each are listed as follows:
\\
\textbf{Learn from Scratch}: FSL methods with a random initialized molecular encoder.
\begin{itemize}
    \item Siamese~\cite{koch2015siamese} uses a duel network to determine if inputs are of the same class.
    \item ProtoNet~\cite{DBLP:conf/nips/SnellSZ17} classifies inputs based on distance between class prototypes.
    \item MAML~\cite{DBLP:conf/icml/FinnAL17} learns a good model parameter initialization and adapts fast on new tasks via optimization.
    \item TPN~\cite{DBLP:conf/iclr/LiuLPKYHY19} constructs a relation graph from input samples via similarity of input and makes label propagation.
    \item EGNN~\cite{DBLP:conf/cvpr/KimKKY19} constructs a relation graph from input samples via similarity of input and predicts labels by edges in a graph.
    \item IterRefLSTM~\cite{altae2017low} adopts a variant of MatchingNet~\cite{DBLP:conf/nips/VinyalsBLKW16} in molecular property prediction.
    \item PAR~\cite{DBLP:conf/nips/WangAYD21} leverages class prototypes to update input representations and designs label propagation for similar inputs in relation graph.
\end{itemize}
\textbf{Leverage Pre-trained Model}: methods which leverage a pretrained molecular encoder~\cite{DBLP:conf/iclr/HuLGZLPL20}.
\begin{itemize}
    \item Pre-GNN~\cite{DBLP:conf/iclr/HuLGZLPL20} is a pretrained GNN model using self-supervised tasks and is directly finetuned on the support set.
    \item Meta-MGNN~\cite{DBLP:conf/www/GuoZYHW0C21} uses Pre-GNN and add self-supervised tasks in meta-training.
    \item Pre-PAR~\cite{DBLP:conf/nips/WangAYD21} is PAR initialized with Pre-GNN.
\end{itemize}
For Tox21, SIDER and MUV, results reported in \cite{DBLP:conf/nips/WangAYD21} are used. {For ToxCast we resplit it and implement baselines using public codes on ToxCast and PCBA.}

\subsection{GS-Meta}
We implement GS-Meta in Pytorch~\cite{DBLP:conf/nips/PaszkeGMLBCKLGA19} and Pytorch Geometric library~\cite{DBLP:journals/corr/abs-1903-02428}.
For a fair comparison we adopt a 5-layer GIN~\cite{DBLP:conf/iclr/XuHLJ19} as molecular encoder for GS-Meta and all baselines. 
The maximum number of optimization step in meta-training is 2000 and meta-testing is evaluated every 100 steps. MLP in Eqn.~\eqref{eq:edge-pred} and $f_{cls}$ in Eqn.~\eqref{eq:classfier} consist two fully connected layers with hidden size 128. Number of candidates $N_{pool}$ and batch size $B$ is 10 and 5 respectively.
Hyper-parameter \textit{k} of connecting molecules is set to 1 in 1-shot and 9 in 10-shot scenario. Table~\ref{tab:hyper-param} illustrates all the hyper-parameters and the results of hyper-parameter sensitivity analysis are in Table~\ref{tab:hyper-sens}. For the number of selected auxiliary properties, on Tox21, SIDER and MUV, we select all possible auxiliary properties in training and testing.
On ToxCast and PCBA, the maximum number is 20 in both training and testing.

\begin{figure}[!t]
\centering
\includegraphics[width=0.99\columnwidth]{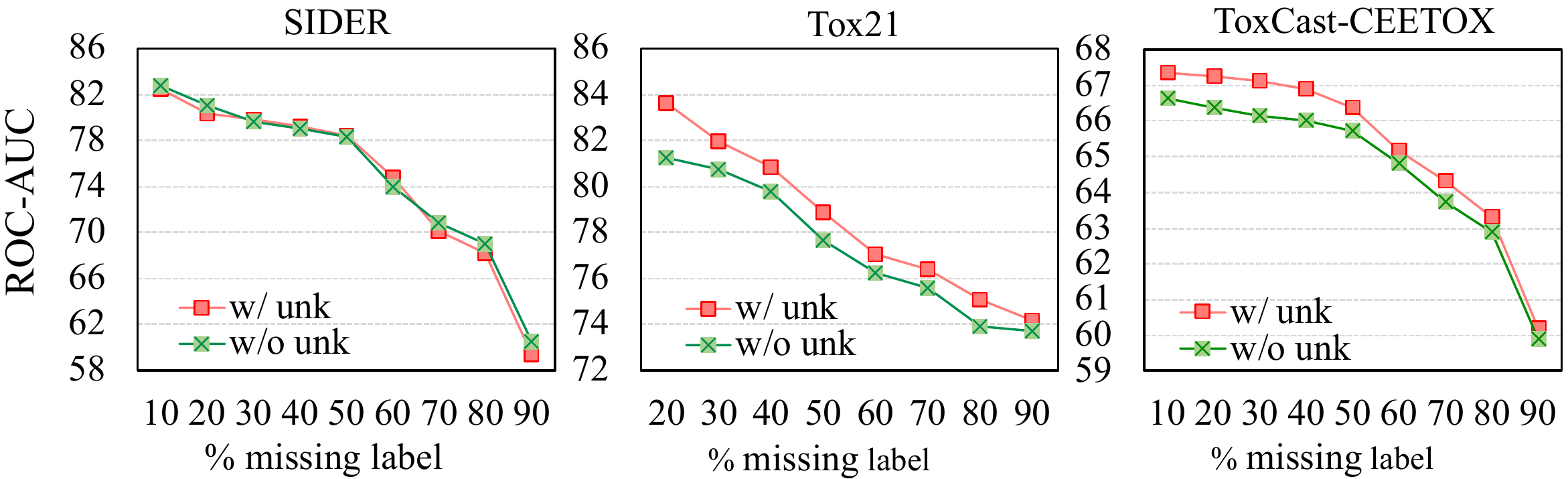} 
\caption{Performance with different missing label ratios in SIDER, Tox21 and ToxCast-CEETOX under the 1-shot scenario.}
\label{fig:missing-1shot}
\end{figure}

\begin{table}[!t]
\centering

\setlength\tabcolsep{12pt}
\scalebox{0.8}{
{
\begin{tabular}{l|cc}
\hline

Method & 1-shot & 10-shot \\
\hline
GS-Meta & \textbf{82.84} & \textbf{83.72} \\
w/o m2m & 82.12($\downarrow$0.72) & 83.09($\downarrow$0.63) \\
w/o E & 57.54($\downarrow$25.30) & 61.85($\downarrow$21.87) \\
w/o S & 81.07($\downarrow$1.78) & 82.23($\downarrow$1.49) \\
w/o CL & 81.36($\downarrow$1.48) & 82.52($\downarrow$1.20) \\
w/o S, w/o CL & 80.69($\downarrow$2.15) & 81.94($\downarrow$1.78) \\
\hline
\end{tabular}
}
}
\caption{Ablation study on SIDER.}
\label{tab:ablation-sider}
\end{table}

\begin{table}[!t]
\centering

\scalebox{0.85}{
{
\begin{tabular}{l|cc}
\hline

Method & 1-shot & 10-shot \\
\hline
PAR & 74.24(0.86s) & 82.74(1.07s) \\
PAR (w/ ATS) & 74.51(2.62s) & 82.66(3.09s) \\
\hline
GS-Meta & \textbf{87.90}(3.04s) & \textbf{88.95}(3.47s) \\
GS-Meta (w/ ATS) & 87.02(6.19s) & 88.24(7.34s)\\
GS-Meta (w/o S) &87.10(2.16s)&88.09(2.48s)\\
\hline
\end{tabular}
}
}
\caption{Performance and time cost on ToxCast-APR.}
\label{tab:scheduler-APR}
\end{table}

\begin{table}[!t]
\centering
\small
\begin{tabular}{ccccc}
\toprule
$\tau$   & 0.01  & 0.1   & 0.5   & 1     \\ \midrule
      & 82.45 & 82.68 & 82.57 & 82.43 \\ \midrule
$\lambda$ & 0.01  & 0.1   & 0.5   & 1     \\ \midrule
      & 82.64 & 82.77 & 82.52 & 81.90 \\ \bottomrule
\end{tabular}
\caption{Hyper-parameter sensitivity analysis on SIDER under 1-shot setting.}
\label{tab:hyper-sens}
\end{table}